\documentclass{article}

\usepackage[preprint]{neurips_2023}

\usepackage{enumerate}
\usepackage{amsmath}
\usepackage{amssymb}
\usepackage{graphicx} % DO NOT CHANGE THIS
\usepackage{multirow}

\usepackage{caption}
\usepackage{bbding} 
\usepackage{stfloats}
\usepackage{cuted}
\usepackage{capt-of}
\usepackage{algorithmicx,algorithm}
\usepackage{color}
\usepackage{colortbl}
\usepackage{xcolor}
\usepackage{subfigure}

\definecolor{blue}{RGB}{0,0,255}
\definecolor{red}{RGB}{255,0,0}
\definecolor{green}{RGB}{0,255,0}
\definecolor{bg}{RGB}{200,200,200}
\definecolor{black}{RGB}{0,0,0}

\usepackage{wrapfig}
\usepackage{listings}

\definecolor{dkgreen}{rgb}{0,0.6,0}
\definecolor{gray}{rgb}{0.5,0.5,0.5}
\definecolor{mauve}{rgb}{0.58,0,0.82}

\usepackage[pagebackref,breaklinks,colorlinks]{hyperref}
\title{ \texttt{Instruct2Act}: Mapping Multi-modality Instructions to Robotic Actions with Large Language Model}

\author{Siyuan Huang$^{1,2}$  \quad Zhengkai Jiang$^4$  \quad Hao Dong$^3$ \quad Yu Qiao$^2$ \quad Peng Gao$^2$ \ Hongsheng Li$^5$\\
$^1$ Shanghai Jiao Tong University,
$^2$ Shanghai AI Laboratory, $^3$ CFCS, School of CS, PKU \\
$^4$ University of Chinese Academy of Sciences,
$^5$ The Chinese University of Hong Kong \\
\{huangsiyuan, gaopeng, qiaoyu\}@pjlab.org.cn, \\ kaikaijiang.jzk@gmail.com, hao.dong@pku.edu.cn, hsli@ee.cuhk.edu.hk}

\begin{document}

\maketitle
\begin{abstract}
Foundation models have made significant strides in various applications, including text-to-image generation, panoptic segmentation, and natural language processing. This paper presents \texttt{Instruct2Act}, a framework that utilizes Large Language Models to map multi-modal instructions to sequential actions for robotic manipulation tasks. Specifically, \texttt{Instruct2Act} employs the LLM model to generate Python programs that constitute a comprehensive perception, planning, and action loop for robotic tasks. In the perception section, pre-defined APIs are used to access multiple foundation models where the Segment Anything Model (SAM) accurately locates candidate objects, and CLIP classifies them. In this way, the framework leverages the expertise of foundation models and robotic abilities to convert complex high-level instructions into precise policy codes. Our approach is adjustable and flexible in accommodating various instruction modalities and input types and catering to specific task demands. We validated the practicality and efficiency of our approach by assessing it on robotic tasks in different scenarios within tabletop manipulation domains. Furthermore, our zero-shot method outperformed many state-of-the-art learning-based policies in several tasks. The code for our proposed approach is available at \href{https://github.com/OpenGVLab/Instruct2Act}{https://github.com/OpenGVLab/Instruct2Act}, serving as a robust benchmark for high-level robotic instruction tasks with assorted modality inputs.
\end{abstract}

\vspace{-0.2cm}
\section{Introduction}
\vspace{-0.2cm}

Recently, Large language models (LLMs) such as GPT-3~\cite{brown2020language}, LLaMA~\cite{touvron2023llama}, and ChatGPT have made unprecedented progress in human-like text generation and understanding of natural language instructions. These models demonstrate remarkable zero-shot generalization abilities after being trained on large collected corpora and absorbing human feedback. The follow-up work Visual ChatGPT~\cite{wu2023visual} incorporates a variety of visual foundation models to achieve visual drawing and editing using a prompt manager. Additionally, VISPROG~\cite{gupta2022visual} proposes a neuro-symbolic approach for complex visual tasks, including image understanding, manipulation, and knowledge retrieval. Inspired by the tremendous potential of combining LLMs and multi-modality foundation models, we aim to develop general robotic manipulation systems. Could we build a ChatGPT-like robotic systems that support robotic manipulation, visual goal-reaching, and visual reasoning?

Developing a general-purpose robotic system capable of executing complex tasks in dynamic environments poses a significant challenge in robotics research. Such a system must possess the ability to perceive the surroundings, select relevant robotic skills, and sequence them accordingly to accomplish long-term goals. Achieving such functionalities requires the integration of various technologies, including perception, planning, and control, to enable the robot to operate autonomously in unstructured environments. Inspired by the strong capability of synthesizing simple Python programs from docstrings of LLMs, CaP~\cite{liang2022code} directly generates the robot-centric policy code based on several in-context example language commands. However, it is restricted to what the perception APIs can provide and struggle to interpret longer and more complex commands due to the high precision requirements of the code.

To address these challenges, we propose a novel approach that utilizes multi-modality foundation models and LLMs to simultaneously implement perceptual recognition, task planning, and low-level control modules. Unlike existing methods such as CaP~\cite{liang2022code}, which directly generates policy codes, we generate decision-making actions that can help reduce the error rate of executing complex tasks. Specifically, we use various foundation models such as SAM and CLIP to accurately locate and classify objects in the environment. We then combine the information with robotic skills to generate decision-making actions by LLMs.

We evaluate our proposed approach on multiple domains and scenarios, including simple object manipulation, visual goal-reaching, and visual reasoning. Our framework provides an easy-to-use general-purpose robotic system and shows strong competitive performance on six representative meta-tasks from VIMABench~\cite{jiang2022vima}. Our proposed approach can serve as a strong baseline method in the field of robotic research and contribute to the development of more intelligent and capable robots.

The contributions of our papers can be summarized as follows:
\vspace{-0.15cm}
\begin{itemize}

\vspace{-0.05cm}  
   \item \textbf{General-purpose robotic system.} We introduce a general-function robotic system, \texttt{Instruct2Act}, that leverages the in-context learning ability of LLMs and multi-modality instructions to generate middle-level decision-making actions from both natural language and visual instructions.
    \vspace{-0.05cm}  
   \item \textbf{Flexible modality inputs.} This paper investigates unified modality instruction inputs on robotic tasks, such as manipulation and reasoning, and presents a flexible retrieval architecture to handle varying instruction types.
    \vspace{-0.05cm}  
   \item  \textbf{Strong zero-shot performance with minimal code overhead.} The proposed \texttt{Instrcut2Act} has shown superior performance in comparison to state-of-the-art learning-based policies, even without fine-tuning. Additionally, the adaptation effect of using foundation models is comparatively minor, as opposed to the training-from-scratch methods.
   
\end{itemize}

\begin{figure}[!t]
    \centering
    \subfigure[Robots are able to execute instructions that are provided as input in natural language.]{\includegraphics[width=0.48\linewidth]{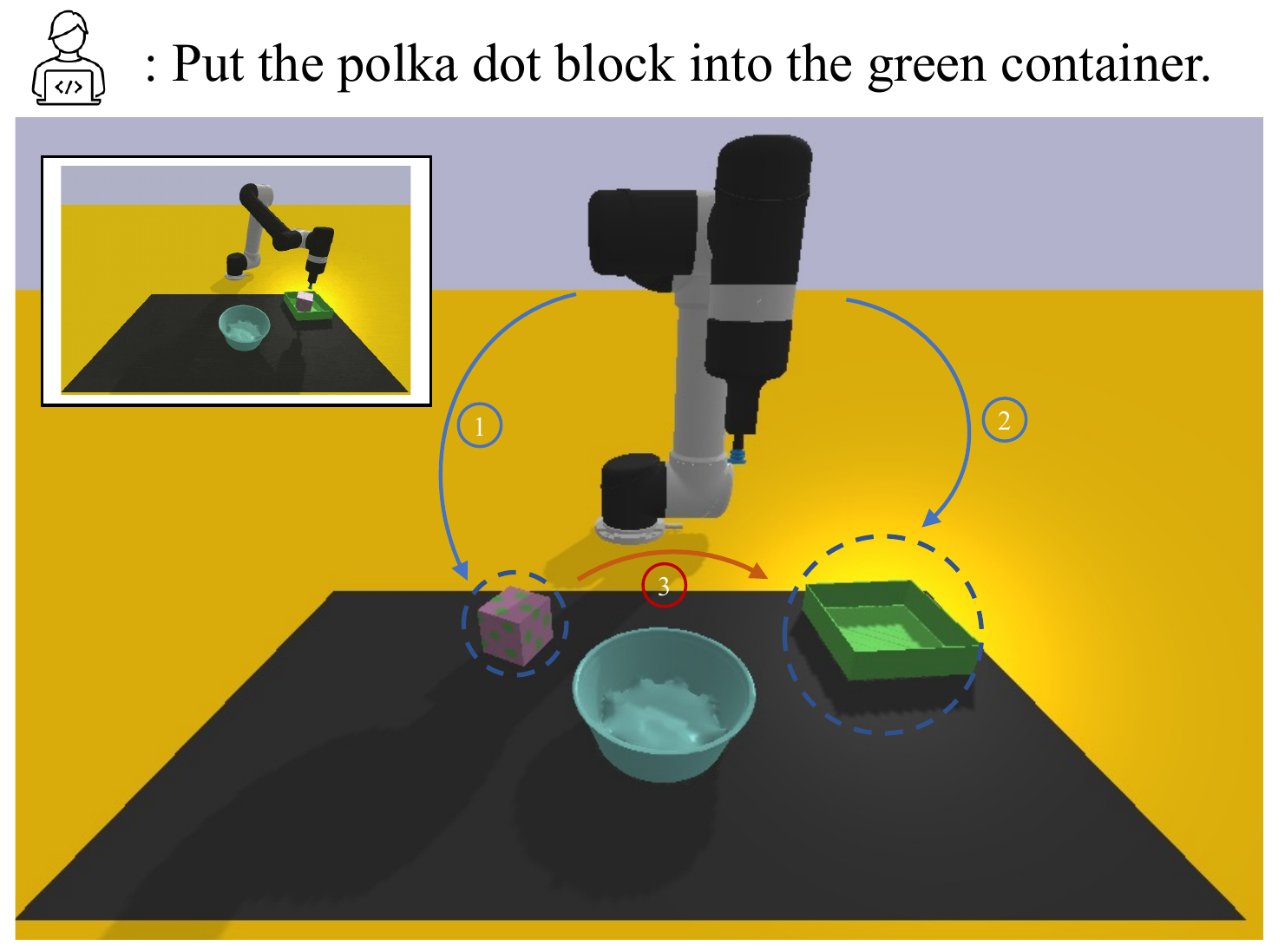}}
    \subfigure[Module examples utilized in \texttt{Instruct2Act}. The modules' definitions are hierarchical and aligned with the robotic system design.]{\includegraphics[width=0.48\linewidth]{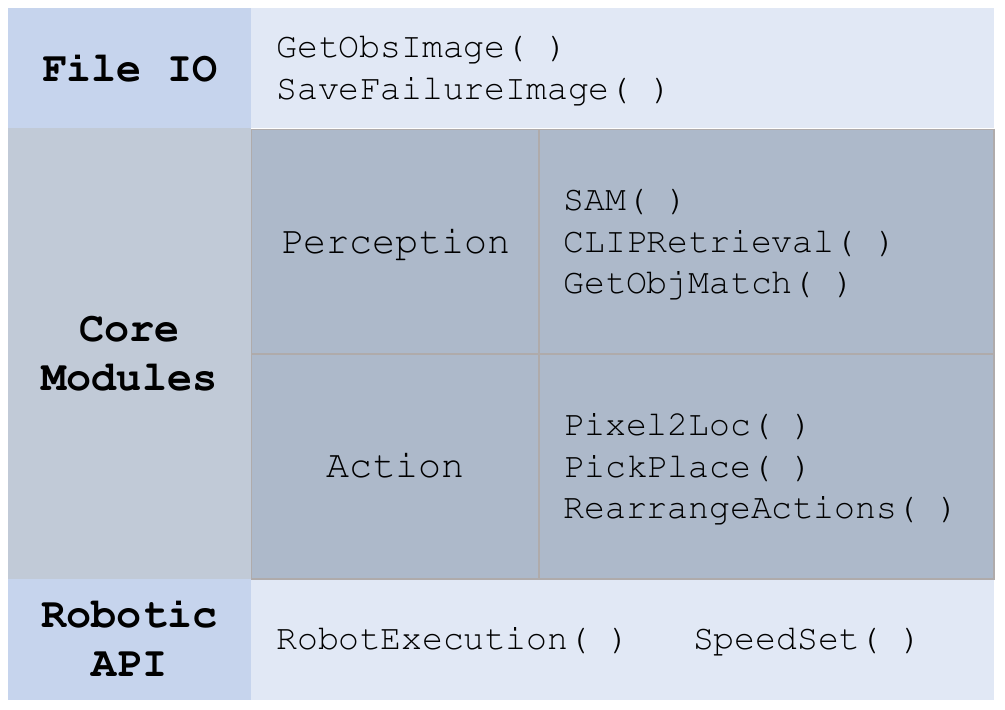}
    \label{fig:modules_api}
    }
    \caption{A robotic task (a) is executed through the invocation of several modules (b)  in \texttt{Instruct2Act}.}
    \label{fig:task_demo_and_api}
\vspace{-0.5cm}
\end{figure}

% \begin{wrapfigure}{l}{0.45\textwidth}
% \vspace{-0.3cm}
% \caption{Robots are able to execute instructions that are provided as input in natural language.}
% \vspace{-0.4cm}
%   \begin{center}
%    \includegraphics[width=0.45\textwidth]{imgs/task_demo.pdf}
%   \end{center}
% \vspace{-0.4cm}
% \end{wrapfigure}

% \begin{wrapfigure}{r}{0.45\textwidth}
% \vspace{-0.3cm}
% \caption{Module examples utilized in \texttt{Instruct2Act}. The module's definition is hierarchical and aligned with the robotic system design.}
% \vspace{-0.4cm}
%   \begin{center}
%    \includegraphics[width=0.45\textwidth]{imgs/modules_api.pdf}
%   \end{center}
%   \label{fig:modules_api}
% \vspace{-0.4cm}
% \end{wrapfigure}

\vspace{-0.2cm}  
\section{Related Works}
\vspace{-0.2cm}  
\label{sec:related_works}

\subsection{Language-driven Robotics} 
\vspace{-0.1cm}
Language in robotics offers not only a user-friendly interface but also the potential for cross-task skill generalization and long-horizon task reasoning. As a result, instruction-based policies have been a popular area of research in robotics~\cite{brohan2023can, shao2021concept2robot, shridhar2018interactive}. Recently, with the emergence of multi-modality models, CLIPORT~\cite{shridhar2022cliport} has given Transporter~\cite{zeng2021transporter} the ability to understand semantics and manipulate objects by encoding text input through CLIP~\cite{radford2021learning}. ~\cite{shridhar2023perceiver} extended the CLIPORT to the 3D domain by employing voxelized observation and action spaces in their Perceiver-Actor model.~\cite{brohan2023can} employed the 540B PaLM~\cite{chowdhery2022palm} to accomplish zero-shot concept grounding in their SayCan model.~\cite{huang2022language} utilized two LLMs in their approach, where one was used for zero-shot planning generation and the other one was used for admissible action mapping.~\cite{huang2022inner} enhanced their method by integrating closed-loop feedback, such as scene descriptors and success detectors, for performing robotic tasks. VIMA~\cite{jiang2022vima} developed a large-scale benchmarking dataset by designing a multimodal prompts-conditioned framework. CaP~\cite{liang2022code} directly generates policy codes with detailed comments and context-specific examples to guide LLM output. ~\cite{cui2023no} conducted a further investigation of the shared autonomy regime. Their approach involves a fusion of the correction signal from human instruction and the static controller with the original policy during inference. PaLM-E\cite{driess2023palm} built a large VL model for embodied agents by integrating the power of 540B PaLM~\cite{chowdhery2022palm} and 22B ViT~\cite{dehghani2023scaling}. Text2Motion~\cite{lin2023text2motion} predicted the goal state and selected feasible actions using LLM while considering geometric constraints. Socratic Models~\cite{zeng2022socratic} generated prompts in their approach by incorporating perceptual information into LLM using VL models. The concurrent work~\cite{wu2023tidybot} uses LLM to summarize the human's preferences given a few examples. Our \texttt{Instruct2Act} framework achieves great flexibility while retaining expert domain knowledge by combining robotic primitive skills with LLM.

\vspace{-0.1cm}
\subsection{Foundation Models on Computer Vision Tasks}
\vspace{-0.1cm}
Several studies~\cite{chen2020uniter,zhang2021vinvl} have utilized frozen pre-trained image encoders to improve the visual features extracted from images. And ~\cite{wang2022internimage} adopted deformable convolutions in their large-scale visual foundation models for better image analysis. In addition, leveraging self-training and a massive dataset of 27M image-text pairs, GLIP~\cite{li2022grounded}achieved strong zero-shot transfer ability. Segment Anything Model (SAM)~\cite{kirillov2023segment}, a segmentation foundation model trained on more than one billion mask samples, allows for zero-shot transfer to diverse tasks through prompt engineering. Furthermore, pre-trained LLMs have  shown significant progress in text understanding and generation~\cite{vaswani2017attention,brown2020language,gao2023llama,touvron2023llama}. And a such breakthrough in LLMs also benefits the VL tasks~\cite{fu2021violet, zhang2021vinvl}. Recently, Recently, there have been explorations to combine the reasoning capacity of LLMs with the visual understanding ability of visual foundation models. VISPROG~\cite{gupta2022visual} utilizes in-context learning in GPT-3 to generate a program for new instruction and demonstrates the system's compositional visual reasoning ability. ViperGPT~\cite{suris2023vipergpt} leverages code-generation models to produce the results of language queries by means of  composing foundation models into subroutines. Visual ChatGPT~\cite{wu2023visual} incorporates multiple visual foundation models and allows users to interact with ChatGPT through the proposed prompt manager, which allows multiple AI models reasoning ability with multi-steps. Similarly, the proposed \texttt{Instruct2Act} aims at endowing robotics the perception ability by incorporating advanced foundation models through the reasoning ability of LLM.

\vspace{-0.1cm}
\subsection{Foundation Models in Robotics}
\vspace{-0.1cm}
In addition to language-conditioned robotic manipulation, the use of foundation models has also led to significant advancements in robotics. LID~\cite{li2022pre} proposes a general approach to sequential decision-making that uses a pre-trained language model (LM) to initialize a policy network where goals and observations are embedded. R3M~\cite{nair2022r3m} explores how visual representations obtained by training on diverse human video data~\cite{grauman2022ego4d} using time-contrastive learning and video-language~\cite{radosavovic2023real} can enable data-efficient learning of downstream robotic manipulation tasks. Meanwhile, CACTI~\cite{mandi2022cacti} suggests a scalable framework for visual imitation learning that utilizes pre-trained models to map pixel values to low-dimensional latent embeddings for improved generalization ability. DALL-E-Bot~\cite{kapelyukh2022dall} uses Stable Diffusion~\cite{rombach2022high} to generate goal scene images that function as guides for robot actions, providing a distinct approach compared to the aforementioned works. In contrast, our \texttt{Instruct2Act} framework employs visual foundation models as modular tools that can be invoked with APIs without the need for fine-tuning, thereby eliminating the need for data collection and training costs.

\vspace{-0.2cm}
\section{Methods}
\vspace{-0.2cm}

Generally, \texttt{Instruct2Act} allows a robot to execute a sequence of actions based on an instruction from the user and an observation image captured by a top-view camera. It aims to change the object state in the environment to match the configuration in the instruction descriptions. \texttt{Instruct2Act} is a language-based robotic system that generates perception-to-action codes using an LLM. Task-related variables, including image crops used in the task instruction and image-to-robot coordinate transformations, are stored in an environment cache $\mathbf{C}$ that can be accessed through an API. The system gains a visual understanding of its manipulation environment by using perception models. Based on this information, the system generates executable action codes that the robot executes with the help of available low-level controllers. To guide the LLM's output, we provide API definitions and in-context examples to the LLM.
 
\begin{figure*}[!t]
\centering
\includegraphics[width=\textwidth]{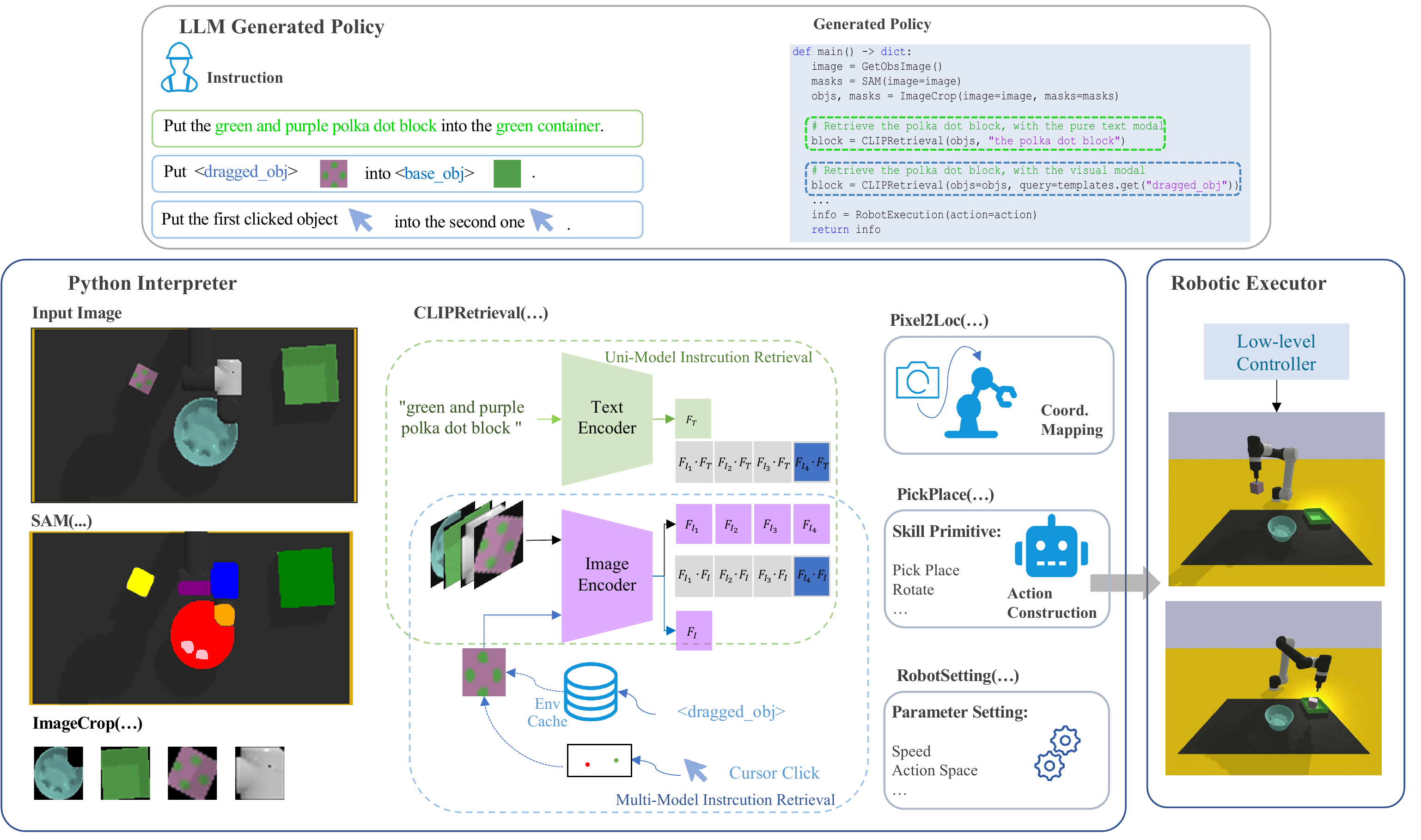}
\caption{The paradigm of our proposed \texttt{Instruct2Act} framework. Starting with the task instruction, the framework utilizes LLM to generate executable code that invokes visual foundation models with APIs to recognize the environment. With recognized object semantic information, we generate plausible actions that are sent to the low-level controller to execute the task. The instructions in green and blue stand for pure-language and multimodal instructions respectively.}
\vspace{-0.3cm}
\label{fig:framework}
\end{figure*}

\vspace{-0.1cm}
\subsection{How to Drive Robotic by LLM}
\vspace{-0.1cm}
\label{sec: llm_as_robotic_driver}

To facilitate LLMs in completing robotic tasks, a designed prompt that guides the LLMs' generation is provided together with specific task instructions. The prompt includes application programming interfaces (APIs) and in-context examples to demonstrate their usage, which are introduced in the following section. The LLM's final input is a sequence of code APIs, usage examples, and task instructions. The LLM's output is a Python function in string format that can be executed by the Python interpreter to drive the robot's action.

Using LLM as the robotic driver has several advantages. Firstly, the robotic policies generated by LLM's API are highly flexible, as in-context examples can be adjusted to guide LLM's behavior and adapt to new tasks. Secondly, by utilizing visual foundation models directly, there is no need to gather training data or to conduct training processes, and any improvements in foundation models can improve action accuracy without incurring additional costs. Lastly, the simple API naming and readable Python code make the generated policy code highly interpretable.

% \noindent \textbf{Object Location Mapping and Action Construction}
After acquiring the input image's semantic information, including the target object's location and semantic class, we need a mapping between the image space and action space to generate executable actions. We use a pre-defined transformation matrix to transfer object location to robot coordinates and apply boundary clamping to prevent unintended actions. The LLM identifies appropriate actions based on instructions and in-context examples.
\vspace{-0.1cm}
\subsection{Prompts for \texttt{Instruct2Act}}
\begin{figure}
  \centering
  \includegraphics[width=0.95\textwidth]{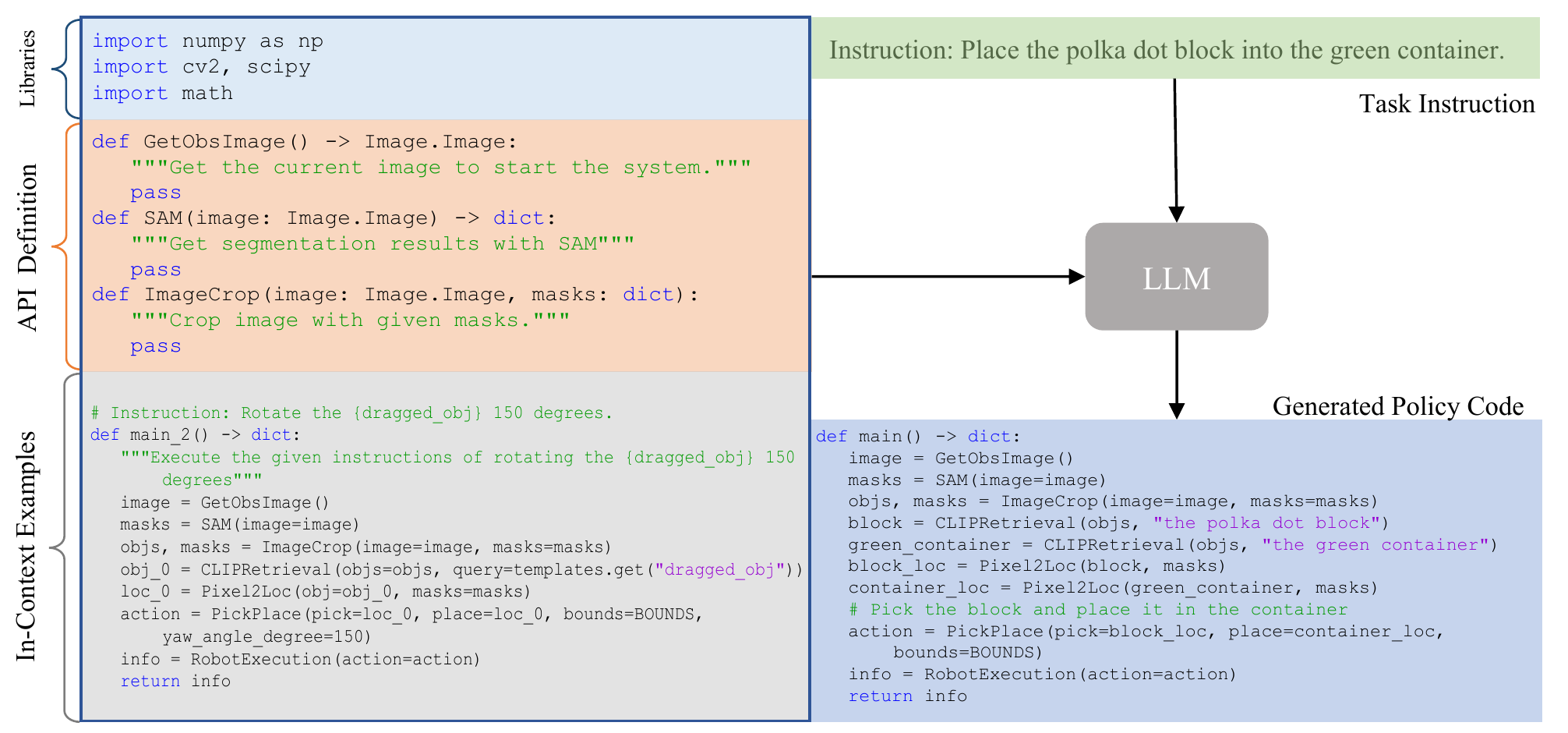}
  \caption{Robotic program generation process. The complete prompt consists of third-party import libraries, API definitions, and in-context examples. }
  \label{fig:prompt_generation}
% \end{wrapfigure}
\vspace{-0.5cm}
\end{figure}

\vspace{-0.1cm}
\label{sec:prompts_and_tools}
Figure~\ref{fig:prompt_generation} illustrates that a complete prompt should include essential information about third-party libraries, API definitions, and in-context examples. The third-party library import information enables the LLM to understand how APIs use the parameter types defined by these libraries to perform calculations and to even create new functions. And we demonstrate its effectiveness in Section~\ref{sec:further_analysis}. We also provide API definitions and descriptions, along with a few in-context examples to demonstrate their usage, similar to~\cite{gupta2022visual}. However, in our approach, we distribute and organize APIs based on robotic system information, as shown in Figure~\ref{fig:modules_api}. Specifically, these APIs are classified according to their functionality within the robot system hierarchy, and this categorized information is provided in the prompt. An example is the \texttt{SAM()} function belonging to \texttt{Perception} module which is the second level core module in the robotic system. So we will add \texttt{\# Second Level: Core Modules} and \texttt{\#\# Perception Modules} before introducing the \texttt{SAM()} API in the prompt. Moreover, unlike ViperGPT~\cite{suris2023vipergpt}, which only offers function-level usage examples, we provide full-logical code examples invoking different modules similarly to the approach of~\cite{gupta2022visual}. This choice is based on the notion that robotic tasks tend to be more intricate yet organized. In contrast to~\cite{gupta2022visual}, we design a prompt that provides coverage for all tasks and has fewer in-context examples. To encourage chain-of-thought reasoning, as done in~\cite{kojima2022large}, we add the prompt \texttt{Think step by step to carry out the instruction} before the inserted specific task instruction. To avoid the LLM from generating too many redundant lines, we explicitly instruct it to only implement the \texttt{main()}  function. Examples of complete prompts are available in Appendix.

\vspace{-0.1cm}
\subsection{Perception with off-the-shelf Foundation Models}
\vspace{-0.1cm}
\label{sec:perception}
% Hs: Put the practice part here and in the implementation details
\texttt{Instruct2Act} leverages off-the-shelf visual foundation models, specifically, the Segment Anything Model (SAM) and CLIP models, accessed by LLM via two designated APIs: \texttt{SAM()} and \texttt{CLIPRetrieval()}. These Python functions load and call the models to conduct the visual analysis. SAM outputs masks for all potential objects in the input image, based on which object crops $I_i$ are extracted correspondingly. These crops are then encoded into $F_{I_{i}}$ using the CLIP image encoder and later utilized for the classification task.  However, pre-trained visual models used directly on downstream tasks without any fine-tuning often suffer from incompleteness or incorrectness. To address these issues caused by the zero-shot paradigm, we insert processing modules between the output of the large model and the downstream tasks. To mitigate the effect of shadows, we apply a gray threshold filter followed by a morphological closing operation to fill up small holes before sending the image to the SAM. After SAM's segmentation operation, we perform a morphological opening operation to eliminate overly small holes or unconnected gaps. We also filter out masks with unreasonable sizes and reduce redundant mask output using Non-Maximum Suppression (NMS). For detailed discussions and visualizations of the processing steps, please refer to Appendix.

\vspace{-0.15cm}
\subsection{Flexible Instruction Modality Manager}
\label{sec:flexible_modality_manager}
\vspace{-0.15cm}
% for pure language, CLIP
% for multi-modal, use language-part to determine task type;
\texttt{Instruct2Act} is flexible and can handle inputs of multiple modalities, such as pure language and language-visual instructions, as depicted in Figure~\ref{fig:framework}. We design a unified retrieval system that utilizes different types of queries to ensure the use of a unified architecture for both types of inputs. 

\vspace{-0.1cm}
\noindent \textbf{Pure-Language Instruction.} For pure language inputs, descriptive sentences are utilized to specify the target object and action. For example, a sentence such as \texttt{Put the green and purple polka dot block into the green container} can be employed. \texttt{Instruct2Act} utilizes the LLM to deduce that the robot needs to fetch the polka dot block from the environment. Therefore, the phrase \texttt{the green and purple polka dot block} acts as the query and is inputted into the CLIP text encoder  to obtain the feature vector $F_{T}$. Finally, the similarity between the query embedding $F_{T}$ and the image crop features $F_{Ii}$ can localize the intended object precisely.

 % single-object: feature matching
% scene-level: feature matching + 匈牙利
\vspace{-0.1cm}
\noindent \textbf{Language-Visual Instruction.}  For the multimodal inputs, the instruction uses an image to describe the target object or the target state. An example instruction is \texttt{Put <$dragged\_obj$> into <$base\_obj$>}. The placeholders in curly braces represent the corresponding images of individual objects, aligning with the LLM model's input format. Given the instruction, the LLM determines the placeholder strings to complete the query which is used to fetch the corresponding object image crop $I$ from the cache $\mathbf{C}$. Then the image crop $I$ is sent to the CLIP image encoder to obtain the feature vector $F_{I}$. We use $F_{I}$ to calculate the similarity with observation image feature vectors $F_{I_{i}}$.

Certain tasks require scene-level understanding, as demonstrated by the task of \texttt{Rearrange to this <scene>.} To fulfill this instruction, we first obtain every possible object and their corresponding feature vectors in the target scene image. We then use the Hungarian algorithm to determine the correspondences between the target scene image and the currently observed images.

\vspace{-0.1cm}
\noindent \textbf{Pointing-Language Enhanced Instruction.} Pointing-language instructions are an effective alternative when the target object cannot be described using pure language instructions and providing image crops is impractical. Specifically, we adopt the cursor movement method from~\cite{2023interngpt} and use cursor clicks to generate point prompts that guide the SAM's segmentation. Additional details can be found in Appendix.

\begin{figure}
    \centering
    \vspace{-0.1cm}
    \includegraphics[width=\textwidth]{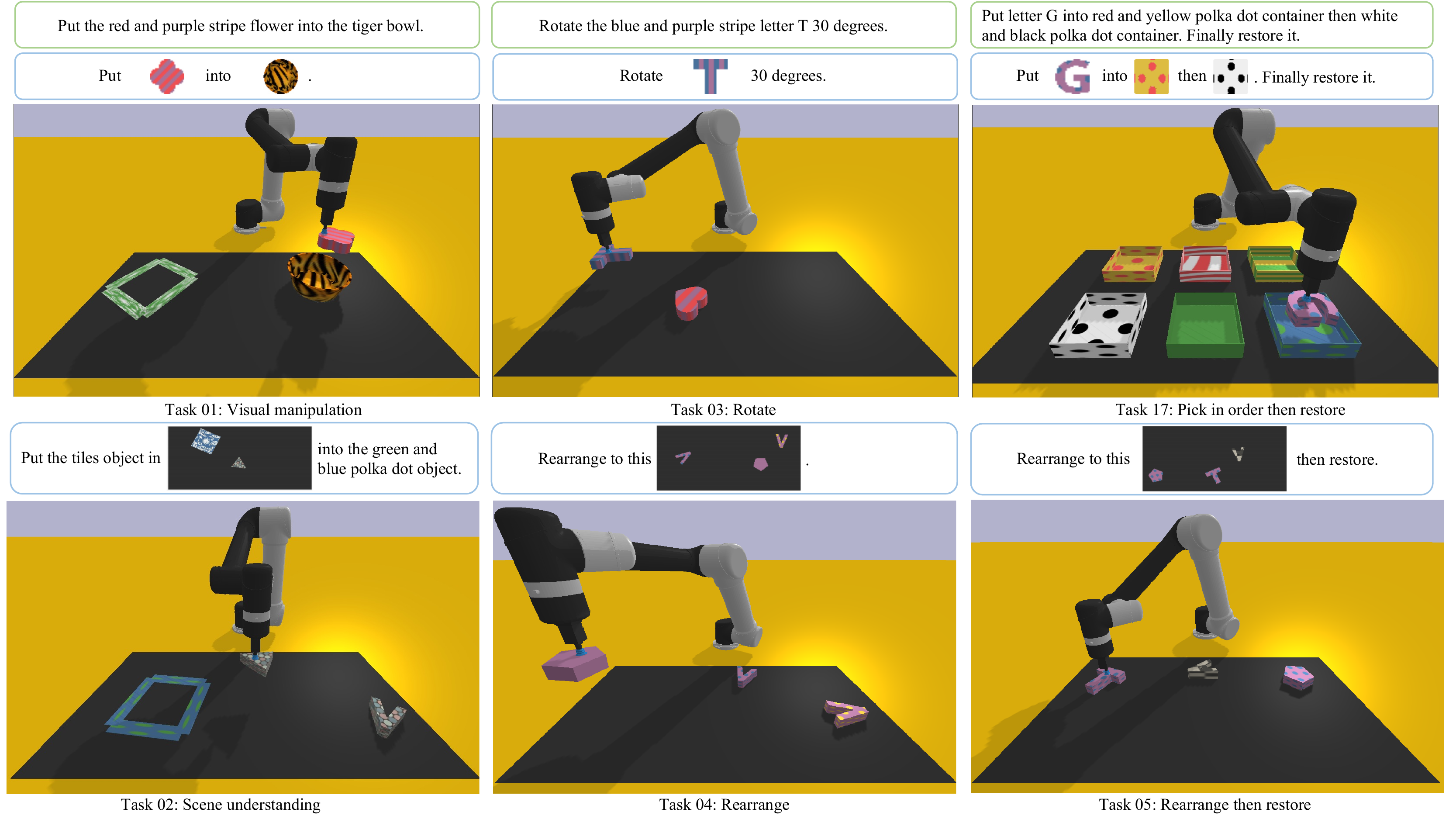}
    \caption{Evaluation task suite. We select six tabletop manipulation meta tasks to evaluate the proposed methods. The instruction in \textcolor{green}{green} and \textcolor{blue}{blue} boxes are uni-modal and multi-modal instructions for the same task respectively.}
    \label{fig:task_suite}
\vspace{-0.6cm}
\end{figure}

\vspace{-0.25cm}
\section{Experiments}
\vspace{-0.25cm}

\label{sec:experiments}
\subsection{Evaluation Task Suite}
% introduces the Vima benchmarking
We select several representative meta tasks from VIMABench~\cite{jiang2022vima} (17 tasks in total), ranging from simple object manipulation to visual reasoning to evaluate the proposed methods in the tabletop manipulation domain, as shown in Fig~\ref{fig:task_suite}.  The evaluation benchmark uses Pybullet~\cite{coumans2016pybullet} as the backend and the default render. In addition to the original multimodal prompt instruction in the VIMABench, we extract object descriptions from the simulator and create a task prompt utilizing natural language that is more commonly used by actual users in their day-to-day lives. Furthermore, the VIMABench contains a 4-level generalization ability evaluation protocol, e.g. L1 placement, L2 combinatorial, L3 novel object, and L4 novel task generalization. And we provide a more detailed task description in Appendix. Each level differs more from the training distribution, and we used the first three levels to evaluate our methods. For more details regarding the evaluation setting, please refer to ~\cite{jiang2022vima}.

\vspace{-0.15cm}
\subsection{Large Language Model}
\vspace{-0.15cm}
\label{sec: llm_exp}
% introduce how to implement
For our experiments, we used two language models: (i) the \textit{text-davinci-003} language model via the OpenAI API, which is a fine-tuned variant of the InstructGPT~\cite{ouyang2022training} language model optimized by using human feedback, and (ii) the LLaMA-Adapter~\cite{gao2023llama, zhang2023llama} via the user interface, which is a lightweight adaptation of the original LLaMA models~\cite{touvron2023llama}. Our approach provided limited prompts to influence the output behavior of the language models without any training or fine-tuning. Although the language models may occasionally generate incomplete or incorrect code with a low rate of errors, such as missing brackets, punctuation, and mismatched cases, the Python Interpreter can detect these errors and inform us to generate new code.

The LLaMA-Adapter language model has a slow inference speed and lacks an API interface due to its local hosting. Therefore, this language model is exclusively used to verify the output. In contrast, the ChatGPT language model is used for large-scale verification experiments as an alternative. A concise comparison of LLaMA-Adapter and ChatGPT is provided in the below experimental section of this paper.

\begin{figure*}[!t]
    \centering
    \vspace{-0.1cm}
    \includegraphics[width=\textwidth]{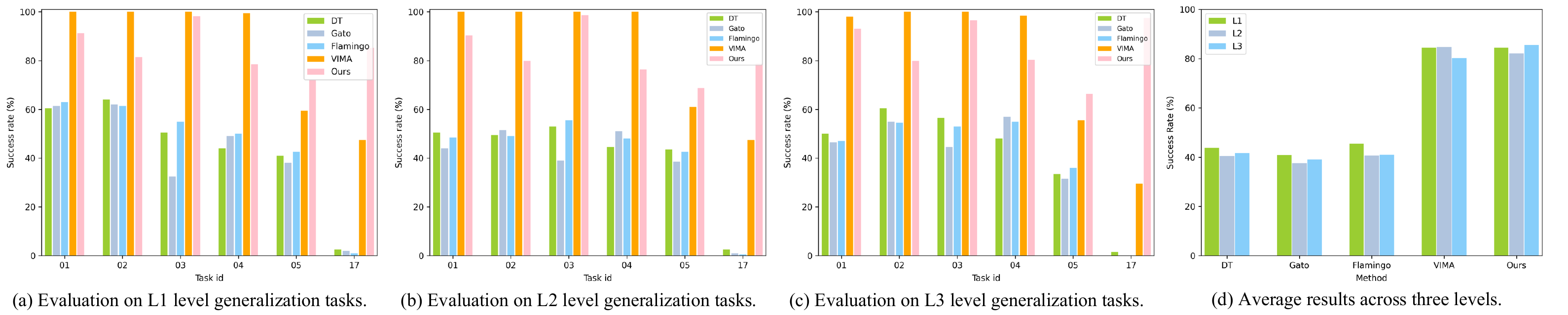}
    \caption{Evaluation results on different generalization levels.}
    \label{fig:evaluation_results_different_level_generalization}
\vspace{-0.3cm}
\end{figure*}

\vspace{-0.2cm}
\subsection{Experiment Results}
\vspace{-0.2cm}
To perform open-vocabulary segmentation, we used open-sourced models such as SAM ViT-H, while for classification, we used CLIP ViT-H-14. Code generation was performed using ChatGPT with the \textit{text-davinci-003} engine. Task success rates were evaluated on 150 instances for each of the six meta-tasks, with three random seeds selected per meta-task to obtain average success rates. The VIMABench simulator determined task success if the final states matched the configurations outlined in the instructions. We conducted our experiments on an NVIDIA 3090Ti GPU. Additionally, we directly used the experiment results of different baselines from~\cite{jiang2022vima}.

Table~\ref{tab:result} presents the experimental results obtained using the proposed \texttt{Instruct2Act} approach on the VIMABench, using the aforementioned foundation models and all processing techniques. The results show that our method achieves comparable average performance with the state-of-the-art (SOTA) learning-based approach, VIMA~\cite{jiang2022vima}, specifically designed for task applications in the multimodal instruction version. Notably, for tasks that require multiple steps to complete, such as Task 05 and Task 17, our \texttt{Instruct2Act} outperforms the previous SOTA by a significant margin, regardless of whether the instructions are single-modal or multimodal. We attribute this improved performance to the strong generalization ability of the large visual model and the powerful reasoning capacity of the LLM. The results also highlight that the performance of multimodal instructions is generally better than that of single-modal instructions. We believe that this is because the former provides a more comprehensive range of information to the model, thereby reducing difficulties for the robot when attempting to reason about the current execution scenario. It is important to note that our method is entirely zero-shot and relies solely on basic task information without using any auxiliary information.

\begin{table*}[!t]
\centering
\begin{tabular}{r|c|c|c|c|c|c|c}
\hline
\rowcolor[HTML]{C0C0C0} 
\multicolumn{1}{c|}{\cellcolor[HTML]{C0C0C0}{\color[HTML]{333333} \textbf{Model}}} & {\color[HTML]{333333} Task 01} & {\color[HTML]{333333} Task 02} & {\color[HTML]{333333} Task 03} & {\color[HTML]{333333} Task 04} &  {\color[HTML]{333333} Task 05}       &  {\color[HTML]{333333} Task 17}      & \multicolumn{1}{l}{\cellcolor[HTML]{C0C0C0}\textbf{Average}} \\ \hline
DT-20M                                                                             & 60.5                           & 64.0                           & 50.5                           & 44.0                           & 41.0          & 2.5           & 43.8                                                         \\ \hline
Gato-20M                                                                           & 61.5                           & 62.0                           & 32.5                           & 49.0                           & 38.0          & 2.0           & 40.8                                                         \\ \hline
Flamingo-20M                                                                       & 63                             & 61.5                           & 55.0                           & 50.0                           & 42.5          & 1.0           & 45.5                                                         \\ \hline
VIMA-20M                                                                           & \textbf{100}                   & \textbf{100}                   & \textbf{100}                   & \textbf{99.5}                  & 59.5          & 47.5          & \textbf{84.4}                                                        \\ \hline
Ours-Multi                                                                         & 91.3                          & 81.4                           & 98.2                           & 78.5                          & \textbf{72.0} & \textbf{85.2} & \textbf{84.4}                                                         \\ \hline
Ours-Single                                                                        & 86.7                           & -                              & 94.6                           & -                              & -             & 63.0          & -                                                            \\ \hline
\end{tabular}
\caption{\textbf{Results on the selected VIMABench tasks for L1 level generalization.} The results are averaged with three seeds per meta-task, and task IDs are identical to the original setting for easy comparison. The numbers in the \textit{Model} column represent the parameter count of the controller, and the instruction prompt modality type is identified as either \textit{Single} or \textit{Multi}.}
\label{tab:result}
\vspace{-0.4cm}
\end{table*}

To further validate the generalization ability, we assessed the effectiveness of our methods on L2 and L3 generalization tasks. The results of different methods are shown in Fig.~\ref{fig:evaluation_results_different_level_generalization}. The results from (a-c) show that our method consistently achieved positive results across all levels and various tasks. Additionally, incorporating foundation models enabled our method to experience less distribution shifting, resulting in minimal performance fluctuations, as shown in Fig.~\ref{fig:evaluation_results_different_level_generalization}-(d).

\vspace{-0.15cm}
\subsection{Further Analysis}
\label{sec:further_analysis}
\vspace{-0.15cm}
\vspace{-0.1cm}
\noindent \textbf{Ablation Studies on Prompt Elements.} We examine the efficacy of each prompt component in the code generation process, including the import information of third-party libraries, API definitions, and contextual examples. Generating executable robotic codes solely based on third-party libraries and without contextual information is nearly impossible. Therefore, in this paper, we consider the import information of third-party libraries as the default input and provide a brief discussion in the following section. For further reference, the complete generated codes is available in Appendix. 

The LLM is able to produce reasonable logic code with detailed explanations by relying solely on API definitions, as demonstrated in Listing~2-4. Unfortunately, in the absence of usage examples, the model can generate unnecessary (actions like \texttt{DistractorActions} and \texttt{RerrangeActions}), as seen in Listing~2, does not return the executed information in Listing~4. Nevertheless, we are pleased to discover that, the LLM can account for execution failures through the invocation of the \texttt{SaveFailureImage} function, which was not present in our original in-context examples.

In contrast, the LLM output results are much more structured and similar to the examples given, when only in-context ones are available.  However, due to the deficiency of functional information, the LLM fails to generate any comments or ambiguously produces incorrect ones, as seen in Listing~5 and Listing~7, where SAM was mistakenly inferred as the abbreviation for Semantic Affinity Module. In addition, variable naming lacks semantic information, and the LLM is unable to add new logic such as failure case handling, beyond the descriptions provided as examples. By providing both the API definitions and in-context examples in the prompt, the LLM can generate accurate and human-readable Python codes, as shown in Listing~8.

After analyzing the observations, we have derived the following conclusions: 1) API style prompts provide more flexibility and enable the LLM to exhibit better reasoning abilities; and 2) the in-context examples style prompts used by the LLM could efficiently sum up and deduce expert information from examples, making them more compatible with structured tasks.

\vspace{-0.1cm}
\noindent \textbf{Ablation Studies on Processing Modules.} Table~\ref{tab:preprocess_ablation} presents ablation studies that validate the effectiveness of the proposed processing modules, including image pre-processing and mask post-processing. The absence of any processing methods leads to significant performance degradation, with only $51\%$ SR,  in line with the findings of the analysis in Section~\ref{sec:perception}. Mask post-processing directly boosts the success rate to $83.0\%$, while further incorporation of image pre-processing achieves the peak success rate of $84.1\%$. It is worth noting that using only image pre-processing leads to no improvement in performance and may even cause degradation in some tasks. This could be attributed to the sub-optimality of the pre-defined parameters, which lack tuning for better generalization.

\begin{table}[!t]
\centering
% \resizebox{0.9\textwidth}{!}{%
\begin{tabular}{cc|c|c|c|c|c|c|c}
\hline
\multicolumn{2}{c|}{Process Method} &
  \cellcolor[HTML]{E0DADA} &
  \cellcolor[HTML]{E0DADA} &
  \cellcolor[HTML]{E0DADA} &
  \cellcolor[HTML]{E0DADA} &
  \cellcolor[HTML]{E0DADA} &
  \cellcolor[HTML]{E0DADA} &
  \cellcolor[HTML]{E0DADA} \\ \cline{1-2}
\multicolumn{1}{c|}{Image} &
  Mask &
  \multirow{-2}{*}{\cellcolor[HTML]{E0DADA}Task 01} &
  \multirow{-2}{*}{\cellcolor[HTML]{E0DADA}Task 02} &
  \multirow{-2}{*}{\cellcolor[HTML]{E0DADA}Task 03} &
  \multirow{-2}{*}{\cellcolor[HTML]{E0DADA}Task 04} &
  \multirow{-2}{*}{\cellcolor[HTML]{E0DADA}Task 05} &
  \multirow{-2}{*}{\cellcolor[HTML]{E0DADA}Task 17} &
  \multirow{-2}{*}{\cellcolor[HTML]{E0DADA}\textbf{Average}} \\ \hline
\multicolumn{1}{c|}{} &  & 70.4          & 34.6          & 88.6          & 41.7          & 15.9          & 54.7          & 51.0          \\ \hline
\multicolumn{1}{c|}{\checkmark} &  & 69.7          & 33.9          & 87.7          & 40.9          & 14.9          & 52.9          & 50.0          \\ \hline
\multicolumn{1}{c|}{} & \checkmark & \textbf{91.7} & 78.2          & 97.4          & 72.9          & \textbf{69.5} & 88.3          & 83.0          \\ \hline
\multicolumn{1}{c|}{\checkmark} & \checkmark & 91.6          & \textbf{80.8} & \textbf{97.8} & \textbf{78.4} & 69.1          & \textbf{87.2} & \textbf{84.1} \\ \hline
\end{tabular}%
% }
\caption{Ablation studies on the effectiveness of the proposed processing modules. The success rates were calculated as the average of performance across three generalized levels.}
\label{tab:preprocess_ablation}
\vspace{-0.3cm}
\end{table}

\vspace{-0.1cm}
\noindent \textbf{Additional Foundational Models.} We experimented with different visual base models to analyze their impact on performance. Specifically, we replaced the original models with SAM-Base and SAM-Large for the semantic segmentation module, and Base-16 and Large-14 for the CLIP model. The results, presented in Table~\ref{tab: sam_ablation} and Table~\ref{tab: clip_ablation}, consistently show that larger foundation models lead to better performance for \texttt{Instruct2Act}. This suggests that our approach can benefit from even stronger foundation models in the future. Additionally, since the visual models are accessed via APIs, they can be easily replaced with other visual foundation models. However, this exploration is currently not our main focus and is left for future work.

\begin{table}
\parbox{.45\linewidth}{
    \centering
\begin{tabular}{r|c|c|c}
\hline
\rowcolor[HTML]{C0C0C0} 
\multicolumn{1}{c|}{\cellcolor[HTML]{C0C0C0}{\color[HTML]{333333} \textbf{}}} & {\color[HTML]{333333} \textbf{Base}} & {\color[HTML]{333333} \textbf{Large}} & {\color[HTML]{333333} \textbf{Huge}} \\ \hline
SR(\%)                                                                        & 75.1                                     & 82.7                                      & 84.1                                    \\ \hline
Parameters(M)                                                                 & 93.7                                     & 312.3                                     & 641.1                                    \\ \hline
\end{tabular}
\caption{Results with different SAM backbones. H-14 is used for the CLIP backbone.}
\label{tab: sam_ablation}
}
\vspace{-0.2cm}
\hfill
\parbox{.45\linewidth}{
    \centering
\begin{tabular}{r|c|c|c}
\hline
\rowcolor[HTML]{C0C0C0} 
\multicolumn{1}{c|}{\cellcolor[HTML]{C0C0C0}{\color[HTML]{333333} \textbf{}}} & {\color[HTML]{333333} \textbf{B-16}} & {\color[HTML]{333333} \textbf{L-14}} & {\color[HTML]{333333} \textbf{H-14}} \\ \hline
SR(\%)                                                                        & 70.0                                 & 76.5                                 & 84.1                                 \\ \hline
Parameters(M)                                                                 & 149.6                                & 427.6                                & 986.1                                \\ \hline
\end{tabular}
\caption{Results with different CLIP backbones. Huge is used for the SAM backbone.}
\label{tab: clip_ablation}}
\vspace{-0.2cm}
\end{table}

\noindent \textbf{Comparison between Different LLMs.} In Section~\ref{sec: llm_exp}, we demonstrate that our proposed \texttt{Instruct2Act} is potentially effective with open-source LLM, in addition to the commercial ChatGPT. Specifically, we use the LLaMA-Adapter~\cite{gao2023llama} and choose Task 01, reporting the success rate on 40 instances in Table~\ref{tab:llama_result}.

% \begin{table}[!t]
\begin{wraptable}{l}{0.55\linewidth}
\vspace{-0.1cm}
% \resizebox{0.95\linewidth} {0.5cm}{
\centering
\begin{tabular}{c|c|c|c}
\hline
\rowcolor[HTML]{C0C0C0} 
Task 01                 & Direct & More Trail & Naive Filtering \\ \hline
\multicolumn{1}{r|}{SR(\%)} & 72.5   & 77.5       & 85.5            \\ \hline
\end{tabular}%
% }
\caption{Results with LLaMA-Adapter.}
\label{tab:llama_result}
\vspace{-0.35cm}
\end{wraptable}
% \end{table}
Remarkably, our \texttt{Instruct2Act} already achieves plausible performance with the LLaMA-Adapter's original output, as shown in Table~\ref{tab:llama_result}. Furthermore, our performance improves to $77.5\%$ when we increase the number of generation trials when the Python interpreter raises an exception. And comparable results can be achieved by using basic filtering, such as re-generating when the environment cache usage is missed in the code.

% \vspace{-0.1cm}
\noindent \textbf{Flexibility and Robustness of \texttt{Instruct2Act}.}
We demonstrate the flexibility and robustness of \texttt{Instruct2Act} by leveraging LLMs' powerful reasoning abilities. Specifically, we evaluate our approach in three scenarios: Human Intervention, Missing Characteristics, and Synonym Replacement. In the Human Intervention scenario, we add sentences to the original instruction to allow human intervention. For example, by appending \texttt{I cancel this task. Stop!}, the LLM can infer that no executable code should be generated. Moreover, the LLM can understand \texttt{That instruction is wrong, use this one.} and generate the correct code for the latter instruction. In the Missing Characteristic scenario, we randomly remove some characteristics or misspell words to test the LLM's grounding ability. Remarkably, the LLM can still ground correctly, demonstrating its robustness. Finally, in the Synonym Replacement scenario, we test the LLM's flexibility by allowing synonym replacement, such as substituting \texttt{Rotate} with \texttt{Spin}. The LLM's flexibility shines as it can handle such variations effortlessly.

% \vspace{-0.1cm}
\noindent \textbf{Extension with The Third-party Library.} As described in Section~\ref{sec:prompts_and_tools}, \texttt{Instruct2Act} can use third-party libraries to perform simple calculations. To demonstrate this, we evaluate the system with the degrees to radians scenario. We offer examples of rotations in degrees for context and inform the system of the \texttt{numpy} library's availability in the prompt, as shown in Fig.~\ref{fig:prompt_generation}. Then, we instruct the system to rotate objects to specific angles in radians, such as 0.5 radians.

\vspace{-0.2cm}
% \definecolor{LightGray}{gray}{0.9}
% \begin{minted}
% [
% frame=lines,
% % framesep=2mm,
% % baselinestretch=1.2,
% % bgcolor=LightGray,
% fontsize=\footnotesize,
% % linenos
% ]
% {python}

% \begin{myblock}

% \textcolor{black}{angle = 0.5 } 
% \textcolor{blue}{ \# in radians} \\
% \textcolor{black}{ \; \, \, yaw\_angle =  angle $*$ 180 $/$ np.pi  } 
% \textcolor{blue}{ \# convert to degrees} 
% \end{myblock}

\begin{lstlisting}
     angle = 0.5 # in radians
     yaw_angle =  angle * 180 / np.pi # convert to degrees
\end{lstlisting}
%  yaw_angle =  angle * 180 / np.pi # convert to degrees
% \end{minted}
%  \vspace{-0.3cm}
\label{tab:prompt_radians}

As demonstrated in the code block above, \texttt{Instruct2Act} uses the provided \texttt{numpy} module to convert degrees to radians before passing the argument to the action function.

% \vspace{-0.1cm}
\noindent \textbf{Limitations.} A notable drawback of \texttt{Instruct2Act} is its high computational cost, as it employs several foundation models to accomplish robotic tasks, which is nearly unacceptable for a real-time robotic system with limited computation resources. Moreover, our method is presently limited by the basic action primitives provided in the chosen VIMABench, such as Pick and Place. However, we are confident that our approach can be readily expanded to include more intricate actions with API extension. Additionally, we have only tested our method in a simulation environment so far, but we plan to investigate real-world applications in the near future. We do not foresee any negative social impact from the proposed work.

\vspace{-0.3cm}
\section{Conclusion} 
\vspace{-0.3cm}

We proposed a \texttt{Instruct2Act} framework to utilize LLM to map multi-modality instructions to sequential actions in the robotics domain. With the LLM-generated policy codes, various visual foundation models  are invoked with APIs to gain a visual understanding of the task sets. To mitigate the gaps in the zero-shot setting, some processing modules are plugged in. Extensive experiments verify that \texttt{Instruct2Act} is effective and flexible in robotic manipulation tasks.

\bibliographystyle{plain}
\bibliography{references}
%\printbibliography

\newpage

\appendix
\section{Appendix}
\label{appendix}

\subsection{Full Prompts in \texttt{Instruct2Act}}
\label{sec:full_prompt}

\begin{lstlisting}[language=Python, xleftmargin=.0\textwidth, xrightmargin=.0\textwidth, caption={An example of a full prompt in \texttt{Instruct2Act}}, label={lst:full_prompt}]
THIRD PARTY TOOLS:
------
You  have access to the following tools:

# Libraries
from PIL import Image
import numpy as np
import scipy
import torch
import cv2
import math
from typing import Union

IMPLEMENTED TOOLS:
------
You have access to the following tools:

# First Level: File IO
templates = {} # dictionary to store and cache the multi-modality instruction
# possible keys in templates: "scene", "dragged_obj", "base_obj"
# NOTE: the word in one instruction inside {} stands for the visual part of the instruction and will be obtained with get operation
# Example: {scene} -> templates.get('scene')
BOUNDS = {} # dictionary to store action space boundary

def GetObsImage(obs) -> Image.Image:
    """Get the current image to start the system.
    Examples:
        image = GetObsImage(obs)
    """
    pass

def SaveFailureImage() -> str:
    """Save images when execution fails
    Examples:
        info = SaveFailureImage()
    """
    pass

# Second Level: Core Modules
## Perception Modules
def SAM(image: Image.Image) -> dict:
    """Get segmentation results with SAM
    Examples:
        masks = SAM(image=image)
    """
    pass

def ImageCrop(image: Image.Image, masks: dict):
    """Crop image with given masks
    Examples:
        objs, masks = ImageCrop(image=image, masks=masks)
    """
    pass

def CLIPRetrieval(objs: list, query: str | Image.Image , pre_obj1: int = None, pre_obj2: int = None) -> np.ndarray:
    """Retrieve the desired object(s) with CLIP, the query could be string or an image
    Examples:
        obj_0 = CLIPRetrieval(objs=objs, query='the yellow and purple polka dot pan') # the query is a string
        obj_0 = CLIPRetrieval(objs=objs, query=templates['dragged_obj']) # the query is image, stored in templates
    """
    pass

def get_objs_match(objs_list1: list, objs_list2: list) -> tuple:
    """Get correspondences of objects between two lists using the Hungarian Algorithm"""
    return (list, list)

## Action Modules
def Pixel2Loc(obj: np.ndarray, masks: np.ndarray) -> np.ndarray:
    """Map masks to specific locations"""
    pass

def PickPlace(pick: np.ndarray, place: np.ndarray, bounds: np.ndarray, yaw_angle_degree: float = None, tool: str = "suction") -> str:
    """Pick and place the object based on given locations and bounds"""
    pass

def DistractorActions(mask_obs: list, obj_list: list, tool: str = "suction") -> list:
    """Remove observed objects that conflict with the goal object list"""
    pass

def RearrangeActions(pick_masks: list, place_masks: list, pick_ind: list, place_ind: list, bounds: np.ndarray, tool: str = "suction") -> list:
    """Composite multiple pick and place actions"""
    pass

# Third Level: Connect to Robotic Hardware
def RobotExecution(action) -> dict
    """Execute the robot, then return the execution result as a dict """
    pass

Examples:
------
Use the following examples to understand tools:
## Example 1
# Instruction: Put the checkerboard round into the yellow and purple polka dot pan.
def main_1() -> dict:
    """Execute the given instructions of placing the checkerboard round into the yellow and purple polka dot pan"""
    image = GetObsImage(obs)
    masks = SAM(image=image)
    objs, masks = ImageCrop(image=image, masks=masks)
    obj_0 = CLIPRetrieval(objs=objs, query='the yellow and purple polka dot pan')
    loc_0 = Pixel2Loc(obj=obj_0, masks=masks)
    obj_1 = CLIPRetrieval(objs=objs, query='the checkerboard round', pre_obj1=obj_0)
    loc_1 = Pixel2Loc(obj=obj_1, masks=masks)
    action = PickPlace(pick=loc_1, place=loc_0, bounds=BOUNDS)
    info = RobotExecution(action=action)
    return info

## Example 2:
# Instruction: Rotate the {dragged_obj} 150 degrees.
def main_2() -> dict:
    """Execute the given instructions of rotating the {dragged_obj} 150 degrees"""
    image = GetObsImage(obs)
    masks = SAM(image=image)
    objs, masks = ImageCrop(image=image, masks=masks)
    obj_0 = CLIPRetrieval(objs=objs, query=templates.get("dragged_obj"))
    loc_0 = Pixel2Loc(obj=obj_0, masks=masks)
    action = PickPlace(pick=loc_0, place=loc_0, bounds=BOUNDS, yaw_angle_degree=150)
    info = RobotExecution(action=action)
    return info

## Example 3
# Instruction: Rearrange to this {scene} then restore.
# Note: for RESTORE operation, direct conduct an inverse operation
def main_3() -> dict:
    """Execute the given instructions of rearranging the objects to match the objects in the given scene"""
    image_obs = GetObsImage(obs)
    image_goal = templates.get("scene")
    masks_obs = SAM(image=image_obs)
    objs_obs, masks_obs = ImageCrop(image=image_obs, masks=masks_obs)
    masks_goal = SAM(image=image_goal)
    objs_goal, masks_goal = ImageCrop(image=image_goal, masks=masks_goal)
    row, col = get_objs_match(objs_list1=objs_goal, objs_list2=objs_obs)
    action_1 = DistractorActions(mask_obs=masks_obs, obj_list=col)
    action_2 = RearrangeActions(pick_masks=masks_obs, place_masks=masks_goal, pick_ind=col, place_ind=row, bounds=BOUNDS)
    action_3 = RearrangeActions(pick_masks=masks_goal, place_masks=masks_obs, pick_ind=row, place_ind=col, bounds=BOUNDS)
    actions = []
    actions.extend(action_1).extend(action_2).extend(action_3)
    info = RobotExecution(action=actions)
    return info


## Example 4
# Instruction: Put the yellow and blue stripe object in {scene} into the orange object.
def main_4() -> dict:
    """Execute the given instructions of placing the yellow and blue stripe object in scene into the orange object"""
    image = GetObsImage(obs)
    masks_obs = SAM(image=image)
    objs_goal, masks_goal = ImageCrop(image=templates['scene'], masks=SAM(image=templates['scene']))
    goal = CLIPRetrieval(objs=objs_goal, query='the yellow and blue stripe object')
    target = CLIPRetrieval(objs=objs_obs, query=objs_goal[goal])
    loc_0 = Pixel2Loc(obj=target, masks=masks_obs)
    obj_1 = CLIPRetrieval(objs=objs_obs, query='the orange object', pre_obj1=target)
    loc_1 = Pixel2Loc(obj=obj_1, masks=masks_obs)
    action = PickPlace(pick=loc_0, place=loc_1, bounds=BOUNDS)
    info = RobotExecution(action=action)
    return info

## Example 5
# Instruction: Put the {dragged_obj} into the {base_obj_1} then {base_obj_2}. Finally restore it into its original container.
def mian_5() -> dict:
    masks = SAM(obs_image)
    objs, masks = ImageCrop(obs_image, masks)
    base_obj_1 = CLIPRetrieval(objs, templates['base_obj_1'])
    base_obj_2 = CLIPRetrieval(objs, templates['base_obj_2'], pre_obj1=base_obj_1)
    dragged_obj = CLIPRetrieval(objs, templates['dragged_obj'], pre_obj1=base_obj_1, pre_obj2=base_obj_2)
    loc_base_obj_1 = Pixel2Loc(base_obj_1, masks)
    loc_base_obj_2 = Pixel2Loc(base_obj_2, masks)
    loc_dragged_obj = Pixel2Loc(dragged_obj, masks)

    action_1 = PickPlace(pick=loc_dragged_obj, place=loc_base_obj_1, bounds=BOUNDS)
    action_2 = PickPlace(pick=loc_base_obj_1, place=loc_base_obj_2, bounds=BOUNDS)
    action_3 = PickPlace(pick=loc_base_obj_2, place=loc_dragged_obj, bounds=BOUNDS)
    actions = [action_1, action_2, action_3]

    info = RobotExecution(action=actions)
    return info

Begin to execute the task:
------
Please solve the following instruction step-by-step. You should implement the main() function and output in the Python-code style.

Instruction: INSERT INSTRUCTION HERE.
\end{lstlisting}

\subsection{Processing Module in \texttt{Instruct2Act}}
\label{sec: app_processing_modules}

When pre-trained models are utilized directly on downstream tasks without any fine-tuning, they inevitably suffer from problems like incompleteness or incorrectness. In light of this, it is advisable to insert processing modules or adapters between the output of the large model and the downstream tasks which can effectively tackle the issues caused by this zero-shot paradigm. The entire loop of execution for robot tasks comprises three modules, namely, perception, planning, and execution. We have curated various processing programs for each of these modules designed for tabletop manipulation domains.

\noindent \textbf{Image Pre-Processing}
In the case of zero-shot SAM outputs, a significant challenge is distinguishing the target object from shadow regions that might appear within the image. As object shadows cannot be grasped, their presence poses additional difficulty for robot grasping tasks. Furthermore, tabletop manipulation domains typically involve camera placement above the robotic arm, leading to large shadows cast onto the operating table. To account for this, we adopted a simple yet efficient image preprocessing methodology: to mitigate the effect of shadows, we employed the gray threshold filter followed by the close morphological operation to fill in small gaps that the filter might produce.

\begin{figure}
    \centering
    \includegraphics[width=0.95\textwidth]{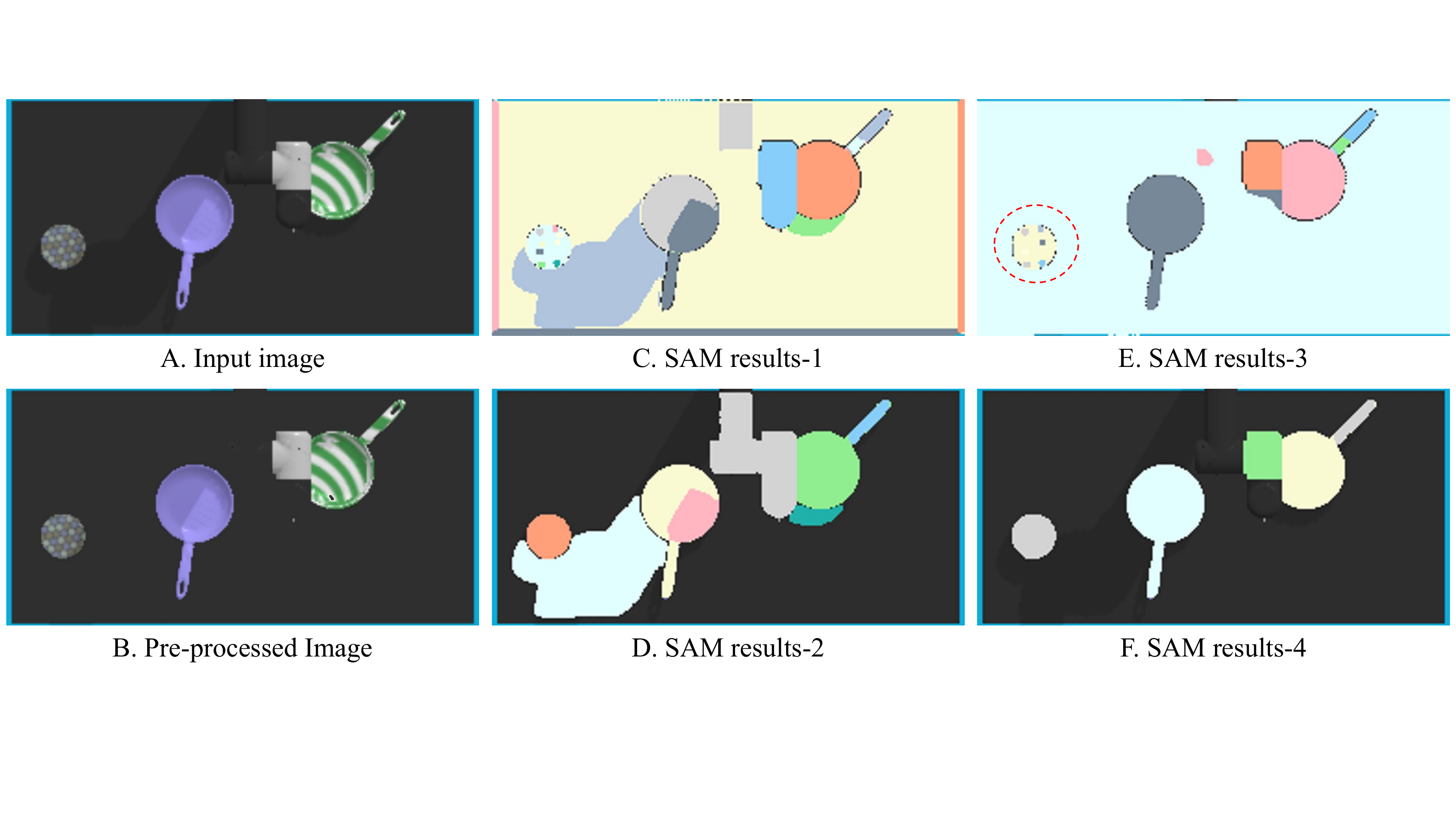}
    \caption{The segmentation outputs w / wo the processing methods. Figure A is the original image which is produced by a top-down camera. Figure B is the image with the proposed image pre-processing pipeline. Figure-	$\left\{C, D, E, F \right \}$ are the SAM results directly on the original image, only with the mask post-processing, only with image pre-processing, and with both processing modules respectively. The portion highlighted by the red circle demonstrates the redundant output of the SAM without additional processing applied to a texture object.}
    \label{fig:seg_result_w_wo_processing}
\end{figure}

\noindent \textbf{Mask Post-Processing} 
In the zero-shot setting, it is observed that the SAM produces multiple segments (e.g. masks) that may be discontinuous or discrete. There may also be detected objects with missing parts, or holes inside, as shown in Fig~\ref{fig:seg_result_w_wo_processing}-C. These misleading semantic mask outputs will inevitably confuse the subsequent modules and greatly challenge the robot grasping task. In order to address the given problem, we have developed a set of processing modules to work with the SAM output. The modules include the following methods:

\begin{itemize}
    \item We apply a filtering process on the output based on the mask's size. This process removes any output that is clearly not part of the target object. Such output may include objects that cannot be moved, like tables or patterns on the target object.

    \item A dilation operation is used to effectively eliminate unrealistic small holes or unconnected gaps. To avoid significant changes to the mask's size due to dilation, we then use an erosion operation. These two procedures combine to create what is known as the opening morphological operation.

    \item In some instances where there are multiple segmentation outputs for a single object, we employ the Non-Maximum Suppression (NMS) operator to reduce redundant mask output.
\end{itemize}

\subsection{Pointing-Language Enhanced Instruction}
\label{sec:app_pointing_language_enhanced}
When utilizing the pointing-language mode of \texttt{Instruct2Act}, the system will display the initial task instruction and observation image. Next, the user will select the target objects by clicking on them with the cursor. These click points will then act as point prompts to guide the SAM's segmentation process. We evaluated this mode on Task 01 and Task 03, averaging the results over 150 instances with one seed.

\begin{table}[]
\centering
\begin{tabular}{l|c|c}
\hline
Instruction Modal       &  Task 01 &  Task  03 \\ \hline
Pointing-Language Enhanced Instruction & 90.7                            & 98.0   \\
\hline
\end{tabular}%
\vspace{0.1cm}
\caption{Results with pointing-language enhanced instruction.}
\label{tab:result_with_pointing_language_enhanced_instruction}
\end{table}

Table~\ref{tab:result_with_pointing_language_enhanced_instruction} shows that our \texttt{Instruct2Act} achieves better results with the pointing-language enhanced mode than with the pure-language instruction mode. This could be due to the stronger prior information provided by the user's click operations.

 \texttt{Instruct2Act}, as discussed in Section \ref{sec:flexible_modality_manager}, can manage instructions with different modalities. For audio instructions, the Whisper model~\cite{radford2022robust} API can be invoked to convert the audio input into text, after which the process described in Section~\ref{sec:flexible_modality_manager} can be followed to guide the LLM and generate policy codes.

\subsection{Evaluation Task Description}
\label{sec:app_task_description}
\noindent \textbf{Simple object manipulation.} The agent is asked to follow the basic instructions to take action. 
\begin{itemize}
    \item Visual Manipulation. The agent is required to pick a specific object and place it into a specified container. The agent needs to first recognize the target objects aligning with the task instruction specified by the natural language description or by the image pattern.  
    
    \item Scene Understanding. The agent must first identify the target object with the described texture by grounding the natural language description and the scene image simultaneously and then put the target object into the container with a specific color.

    \item Rotation. The agent is required to rotate a specific object by certain degrees along the z-axis. 
\end{itemize}

\noindent \textbf{Visual goal-reaching.} The agent is asked to manipulate the objects to match the goal states described by the goal scene image,
\begin{itemize}
    \item Rearrange. The agent is required to rearrange the target objects to reach the goal configuration. The agent needs to identify the possible existing distractors and move them away to avoid position conflicts. 

    \item  Rearrange then restore. The agent is required to restore the object placements after the rearrangement operations.
\end{itemize}

\noindent \textbf{Visual Reasoning} The agent is asked to make decisions and take actions where reasoning and memory ability are required. 
\begin{itemize}
    \item Pick in order then restore. The agent is required to pick and place the target object sequentially into different containers and finally restore it to the initial container.   
\end{itemize}

VIMABench~\cite{jiang2022vima} presents a 4-level evaluation protocol that progressively increases in difficulty for trained agents. Our experiments utilize the first 3 levels of this protocol, which test the generalization abilities of our agents. Level 1 (L1) placement generalization randomly arranges the placement of target objects, whereas L2 combinatorial generalization generates new combinations of target materials and object descriptions. Finally, L3 novel object generalization tests our agents' ability to generalize to novel materials and objects.

\subsection{Ablation Studies on Prompt Element}
\label{sec:ablation_studies_on_prompt_element}

We use the same task instruction \texttt{\textit{Put the polka dot block into the green container.}} for all experiments here. Because the outputs of LLMs are somewhat random, we will display the results of three consecutive outputs.

\begin{lstlisting}[language=Python,  xleftmargin=.0\textwidth, xrightmargin=.0\textwidth, caption={Generation with Only API Definition provided in Prompt. First Trial}, label={lst:only_api_trial_1}]
def main(): 
    # Step 1: Get the current observation image 
    image = GetObsImage(obs)
    
    # Step 2: Segment the image to get masks
    masks = SAM(image=image)
    
    # Step 3: Crop the image based on masks
    objs, masks = ImageCrop(image=image, masks=masks)
    
    # Step 4: Retrieve the object(s) of interest using CLIP
    query = 'polka dot block' # define the query
    polka_dot_block = CLIPRetrieval(objs=objs, query=query)
    
    # Step 5: Map masks to specific object locations
    obj_loc = Pixel2Loc(obj=polka_dot_block, masks=masks)
    
    # Step 6: Get the pick and place locations and action boundaries
    pick_loc = obj_loc # we want to pick up the polka dot block
    place_loc = BOUNDS['green_container'] # the green container is the place location
    bounds = BOUNDS['PickPlace_bounds'] # limits of the pick and place action
    
    # Step 7: Perform any necessary distracting actions
    obj_list = ['polka dot block'] # define the goal object list 
    masks_obs = np.delete(masks, polka_dot_block, axis=0) # remove the polka dot block from the observation masks
    DistractorActions = DistractorActions(mask_obs=masks_obs, obj_list=obj_list)
    
    # Step 8: Plan and execute the pick and place action
    pick_masks = masks[polka_dot_block]
    pick_ind = [0] # We only want to pick up one object
    place_masks = np.expand_dims(place_loc, axis=0) # expand dims to match the pick loc array
    place_ind = [0] # We only have one object to place
    action_seq = RearrangeActions(pick_masks=pick_masks, place_masks=place_masks, pick_ind=pick_ind, place_ind=place_ind, bounds=bounds)
    action_seq = DistractorActions + action_seq # combine distractor actions with pick and place actions
    for action in action_seq:
        result = RobotExecution(action)
        time.sleep(result['time_elapsed']) # wait for the robot to finish the action
    return result
\end{lstlisting}

\begin{lstlisting}[language=Python,  xleftmargin=.0\textwidth, xrightmargin=.0\textwidth, caption={Generation with Only API Definition provided in Prompt. Second Trial}, label={lst:only_api_trial_2}]
def main(): 
    # Step 1: Get the current observation image
    obs_image = GetObsImage(obs)
    
    # Step 2: Use SAM to get segmentation results
    masks = SAM(image=obs_image)
    
    # Step 3: Crop the image to get object bounding boxes
    objs, masks = ImageCrop(image=obs_image, masks=masks)
    
    # Step 4: Use CLIP to retrieve the polka dot block
    polka_dot = CLIPRetrieval(objs=objs, query='polka dot block')
    
    # Step 5: Use CLIP to retrieve the green container
    green_container = CLIPRetrieval(objs=objs, query='green container')
    
    # Step 6: Map masks to specific locations
    polka_dot_loc = Pixel2Loc(obj=polka_dot, masks=masks)
    green_container_loc = Pixel2Loc(obj=green_container, masks=masks)
    
    # Step 7: Pick and place the polka dot into the green container
    action = PickPlace(pick=polka_dot_loc, place=green_container_loc, bounds=BOUNDS)
    
    # Step 8: Execute the action using the robotic hardware
    result = RobotExecution(action)
    
    # Step 9: Save images if execution fails
    if not result['success']:
        SaveFailureImage()
    
    # Step 10: return the result
    return result
\end{lstlisting}

\begin{lstlisting}[language=Python,  xleftmargin=.0\textwidth, xrightmargin=.0\textwidth, caption={Generation with Only API Definition provided in Prompt. Third Trial}, label={lst:only_api_trial_3}]
def main(): 
    # Step 1: Get the current observation image
    obs_image = GetObsImage(obs)
    
    # Step 2: Use SAM to get segmentation results
    masks = SAM(image=obs_image)
    
    # Step 3: Crop the image to get object bounding boxes
    objs, masks = ImageCrop(image=obs_image, masks=masks)
    
    # Step 4: Use CLIP to retrieve the polka dot block
    polka_dot = CLIPRetrieval(objs=objs, query='polka dot block')
    
    # Step 5: Use CLIP to retrieve the green container
    green_container = CLIPRetrieval(objs=objs, query='green container')
    
    # Step 6: Map masks to specific locations
    polka_dot_loc = Pixel2Loc(obj=polka_dot, masks=masks)
    green_container_loc = Pixel2Loc(obj=green_container, masks=masks)
    
    # Step 7: Pick and place the polka dot into the green container
    action = PickPlace(pick=polka_dot_loc, place=green_container_loc, bounds=BOUNDS)
    
    # Step 8: Execute the action using the robotic hardware
    result = RobotExecution(action)
    
    # Step 9: Save images if execution fails
    if not result['success']:
        SaveFailureImage()
    
    # Step 10: Print the result
    print(result)
\end{lstlisting}

\begin{lstlisting}[language=Python,  xleftmargin=.0\textwidth, xrightmargin=.0\textwidth, caption={Generation with Only In-Context Examples provided in Prompt. First Trial}, label={lst:only_examples_trial_1}]
def main() -> dict:
    """Execute the given instructions of placing the polka dot block into the green container"""
    image = GetObsImage(obs)
    masks = SAM(image=image)
    objs, masks = ImageCrop(image=image, masks=masks)
    obj_0 = CLIPRetrieval(objs=objs, query="the polka dot block")
    loc_0 = Pixel2Loc(obj=obj_0, masks=masks)
    obj_1 = CLIPRetrieval(objs=objs, query="the green container")
    loc_1 = Pixel2Loc(obj=obj_1, masks=masks)
    action = PickPlace(pick=loc_0, place=loc_1, bounds=BOUNDS)
    info = RobotExecution(action=action)
    return info
\end{lstlisting}

\begin{lstlisting}[language=Python,  xleftmargin=.0\textwidth, xrightmargin=.0\textwidth, caption={Generation with Only In-Context Examples provided in Prompt. Second Trial}, label={lst:only_examples_trial_2}]
def main() -> dict:
    # Use object detection to get the object and container locations from the RGB image.
    image = GetObsImage(obs)
    masks = SAM(image=image)
    objs, masks = ImageCrop(image=image, masks=masks)
    block_obj = CLIPRetrieval(objs=objs, query='the polka dot block')
    container_obj = CLIPRetrieval(objs=objs, query='the green container')

    # Convert the pixel locations of the objects into robot reference frame (in meters).
    block_loc = Pixel2Loc(obj=block_obj, masks=masks)
    container_loc = Pixel2Loc(obj=container_obj, masks=masks)

    # Move the robot's arm to pick and place the block into the container.
    action = PickPlace(pick=block_loc, place=container_loc, bounds=BOUNDS)
    info = RobotExecution(action=action)
    return info
\end{lstlisting}

\begin{lstlisting}[language=Python,  xleftmargin=.0\textwidth, xrightmargin=.0\textwidth, caption={Generation with Only In-Context Examples provided in Prompt. Third Trial}, label={lst:only_examples_trial_3}]
def main() -> dict:
    """Execute the given instructions of placing the polka dot block into the green container"""
    # Obtain the image of the scene
    image = GetObsImage(obs)

    # Segment the image using Semantic Affinity Module (SAM)
    masks = SAM(image=image)

    # Crop image to obtain individual objects and their masks
    objs, masks = ImageCrop(image=image, masks=masks)

    # Retrieve the polka dot block object using OpenAI's CLIP model
    obj_0 = CLIPRetrieval(objs=objs, query='the polka dot block')

    # Get the location of the green container
    obj_1 = CLIPRetrieval(objs=objs, query='the green container')
    loc_1 = Pixel2Loc(obj=obj_1, masks=masks)

    # Move the polka dot block from its current location to the green container
    loc_0 = Pixel2Loc(obj=obj_0, masks=masks)
    action = PickPlace(pick=loc_0, place=loc_1, bounds=BOUNDS)
    info = RobotExecution(action=action)

    # Return the execution information
    return info
\end{lstlisting}

\begin{lstlisting}[language=Python,  xleftmargin=.0\textwidth, xrightmargin=.0\textwidth, caption={Generation with both API Definition and In-context Examples provided in Prompt. First Trial}, label={lst:all_trial_1}]
def main() -> dict:
    # first, capture an image of the current scene
    image = GetObsImage(obs)

    # generate object masks using the SAM module
    masks = SAM(image=image)

    # identify the objects in the scene using the masks
    objs, masks = ImageCrop(image=image, masks=masks)

    # use CLIPRetrieval to find the polka dot block and the green container
    polka_dot_block = CLIPRetrieval(objs=objs, query="the polka dot block")
    green_container = CLIPRetrieval(objs=objs, query="the green container")

    # convert the object masks to location coordinates
    polka_dot_block_loc = Pixel2Loc(obj=polka_dot_block, masks=masks)
    green_container_loc = Pixel2Loc(obj=green_container, masks=masks)

    # use the PickPlace function to pick up the polka dot block and place it in the green container
    action = PickPlace(pick=polka_dot_block_loc, place=green_container_loc, bounds=BOUNDS)

    # execute the action on the robot using the RobotExecution function and return the result
    info = RobotExecution(action=action)
    return info
\end{lstlisting}

\end{document}